\documentclass{article}

     \PassOptionsToPackage{numbers, compress}{natbib}

\usepackage[preprint]{neurips_2022}

\usepackage[utf8]{inputenc} 
\usepackage[T1]{fontenc}    
\usepackage{hyperref}       
\usepackage{url}            
\usepackage{booktabs}       
\usepackage{amsfonts}       
\usepackage{nicefrac}       
\usepackage{microtype}      
\usepackage{xcolor}         
\usepackage{graphicx}       

\DeclareSymbolFont{extraup}{U}{zavm}{m}{n}
\DeclareMathSymbol{\varheart}{\mathalpha}{extraup}{86}
\DeclareMathSymbol{\vardiamond}{\mathalpha}{extraup}{87}

\title{SGPT: GPT Sentence Embeddings for Semantic Search}

\author{
  Niklas Muennighoff \\
  Peking University \\
  \texttt{muennighoff@stu.pku.edu.cn} \\
}

\begin{document}

\maketitle

\begin{abstract}

Decoder transformers have continued increasing in scale reaching hundreds of billions of parameters. Due to their scale the same decoder sets state-of-the-art results on various language tasks via prompting or fine-tuning. Yet, these large foundation models remain unusable for the related fields of semantic search and sentence embeddings. This prevents possibly new state-of-the-art results and forces organizations to train and maintain separate models. To this end, we propose SGPT to use decoders for sentence embeddings and semantic search via prompting or fine-tuning. At 5.8 billion parameters SGPT improves on the previously best sentence embeddings by a margin of 7\% and outperforms a concurrent method with 175 billion parameters as measured on the BEIR search benchmark. Code, models and result files are freely available at \url{https://github.com/Muennighoff/sgpt}.

\end{abstract}

\section{Introduction}

Semantic search consists of two parts: \textit{Search} refers to finding the top $k$ answers from a document corpus given a query. \textit{Semantic} refers to understanding the documents and queries beyond keywords. Transformers \citep{vaswani2017attention} are the dominant semantic architecture \cite{cer2018universal, thakur2021beir} competing with non-semantic models like BM25 \cite{robertson2009probabilistic}. Search applications like Google \cite{nayak2021google} or Bing \cite{zhu2021bing} rely on transformers to provide semantically relevant results. However, they have been limited to BERT-like encoder-only transformers \cite{nayak2021google, zhu2021bing, devlin2018bert, reimers2019sentence, formal2021splade, ni2021large}.

Meanwhile, GPT-like decoder-only transformers \cite{radford2018improving} have been the focus of recent scaling efforts of up to 540 billion parameters \cite{chowdhery2022palm}. Increasing language model parameters has been repeatedly shown to improve downstream zero-shot and fine-tuning performance on a variety of language tasks \cite{brown2020language, rae2021scaling, chowdhery2022palm}. For example, increasing scale has allowed decoder-only transformers to outperform all encoder-only and catch-up with encoder-decoder transformers on the SuperGLUE benchmark \cite{wang2019superglue, chowdhery2022palm}.

However, the related fields of semantic search and language embeddings have not been part of the proliferation of decoders. They are dominated by comparatively small encoders \cite{thakur2021beir}, as it remains unclear how to extract semantically meaningful embeddings from decoders and use them for semantic search. Methods to do so are desirable for two reasons: 

\paragraph{Performance}
Taking advantage of the available scale of decoders has the potential to produce new state-of-the-art results in search. Available encoders are orders of magnitude smaller \cite{devlin2018bert, liu2019roberta, raffel2019exploring}. Google search, for example, processes an estimated 4 billion searches daily \cite{googlestats}, thus better search models could have wide-reaching impacts.

\paragraph{Compute Savings} 
Large-scale pre-trained decoders have been reused for different tasks via prompting or fine-tuning \cite{brown2020language, rae2021scaling, chowdhery2022palm}. A well-performing method to extract embeddings from billion parameter decoders may prevent the need to train and maintain separate encoder and decoder models. Training just one large decoder and reusing it for search prevents additional cost to the environment \cite{bommasani2021opportunities}.

\begin{figure*}[t]
    \centering
    \begin{center}
        \includegraphics[width=\textwidth]{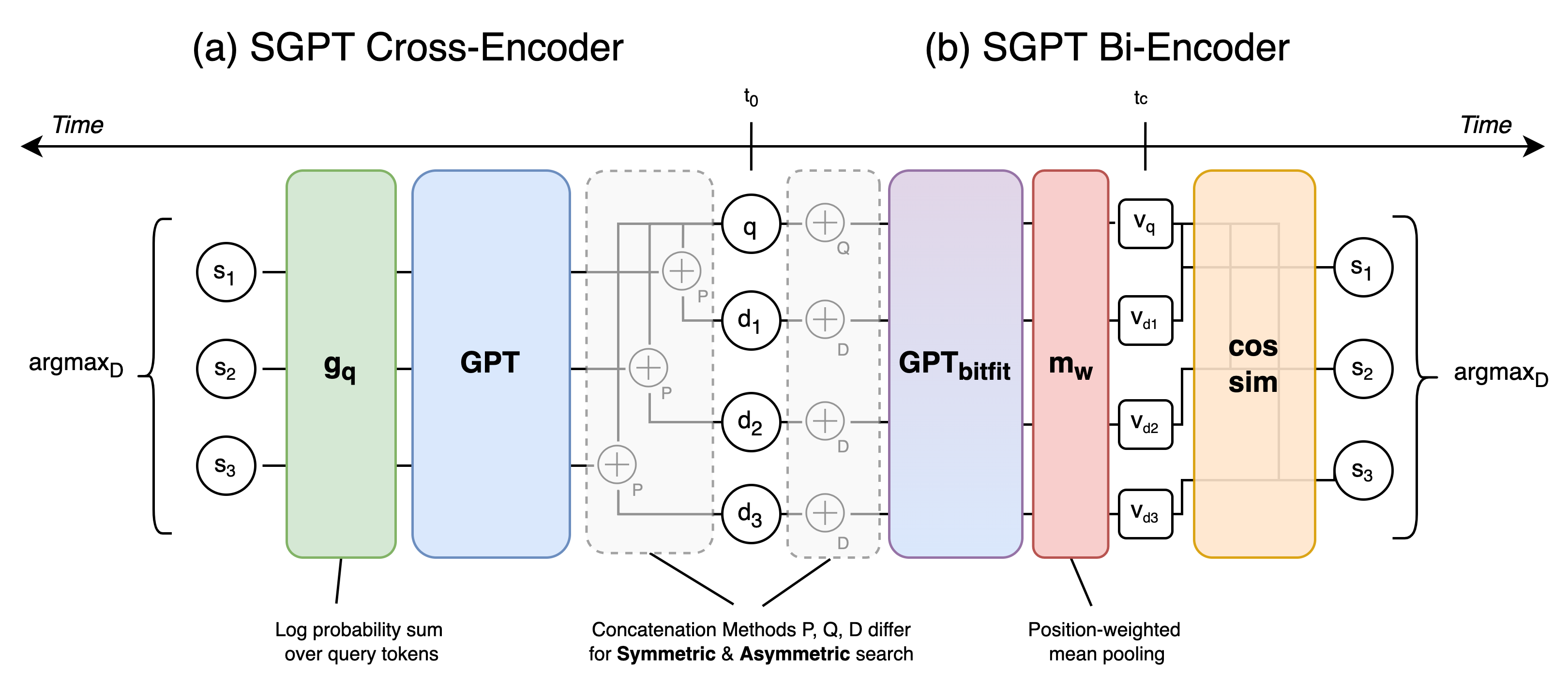}
        \caption{Given a query $q$, documents $d_{1-3}$, SGPT ranks the documents with scores $s_{1-3}$. (a) The Cross-Encoder concatenates queries and documents and encodes them together. Scores are extracted log probabilities. (b) The Bi-Encoder separately encodes queries and documents. Resulting document vectors $v_{1-3}$ can be cached and retrieved at time $t_c$, when a new query comes in. Scores are cosine similarities.}
        \label{fig:sgptgraphic}
    \end{center}
\end{figure*}

In this work, we propose SGPT to apply decoder-only transformers to semantic search and extract meaningful sentence embeddings from them. We distinguish four settings: Cross-Encoder vs Bi-Encoder, Symmetric vs Asymmetric. See Figure \ref{fig:sgptgraphic} and \S\ref{sec:background}.

In the Bi-Encoder setting, we propose SGPT-BE using position-weighted mean pooling and contrastive fine-tuning of only bias tensors (BitFit \cite{zaken2021bitfit}). We show that BitFit is competitive with full fine-tuning performance for both encoders (SBERT) \cite{reimers2019sentence} and decoders (SGPT) despite changing <0.1\% of pre-trained parameters. When controlling for size, our decoders closely trail the performance of encoders. When scaling up, SGPT-BE-5.8B sets state-of-the-art results on BEIR and USEB for asymmetric and symmetric search.

In the Cross-Encoder setting, we propose SGPT-CE using log probability extraction of pre-trained GPT models. The method is applicable to symmetric or asymmetric search by changing the prompt. At scale, the model sets an unsupervised state-of-the-art on BEIR.

In summary, our contributions are three-fold:
\begin{itemize}
    \item For SGPT-BE in \S\ref{sec:be}, we develop a new pooling method and show the usefulness of bias-only fine-tuning for embeddings. At 5.8B parameters, it produces the best natural language embeddings available by a margin of 7\% for the example of semantic search.
    \item For SGPT-CE in \S\ref{sec:ce}, we show how to use GPT for search via log probabilities without fine-tuning. At 6.1B parameters, it has the best unsupervised performance on BEIR by a margin of 8\%.
    \item We provide free, more performant alternatives to commonly used endpoints, such as OpenAI's Search and Similarity Embeddings and the OpenAI Search endpoint available at \url{https://github.com/Muennighoff/sgpt}.
\end{itemize}






\section{Related Work}\label{sec:background}

In this section, we explain two dimensions fundamental to our work: Cross-Encoders vs Bi-Encoders and Symmetric vs Asymmetric Search. We highlight work in those areas relevant to ours.

\textbf{Cross-Encoders} encode query and document at the same time. BERT \cite{devlin2018bert} is used as a Cross-Encoder by separating the query from the document with a $[SEP]$ token. They are then passed through the transformer network together. Each new query requires $k$ forward passes given a corpus of $k$ documents. There is no existing literature on using GPT models as Cross-Encoders, however, we suspect the OpenAI Search API \cite{openaisearch} uses a GPT-based Cross-Encoder. We include results based on querying their API, as well as a BERT-based state-of-the-art Cross-Encoder in our benchmarks.

\textbf{Bi-Encoders} encode query and document separately. SBERT \cite{reimers2019sentence} extends BERT to the Bi-Encoder setting via supervised fine-tuning and a pooling operation across the sequence output. The resulting document vectors can be cached. A new query requires only one forward pass through the transformer to produce the query vector. The query vector can then be scored against the cached document vectors with a similarity function. Embeddings from Bi-Encoders can be used for non-search tasks such as clustering or as input features of machine learning models \cite{reimers2019classification}. While non-semantic models like keyword-based BM25 \cite{robertson2009probabilistic} remain extensively used, the field increasingly focuses on neural models using transformers \cite{vaswani2017attention}. Contriever \cite{izacard2021towards} shows the utility of unsupervised contrastive training for pre-trained encoders. Their best model that we compare with also adds supervised fine-tuning as a third training stage. GTR \cite{ni2021large} explores the effect of scaling up encoders on semantic search tasks also in a three-stage training setup. They find that despite keeping the embedding size fixed, more parameters increase the performance of encoders. Usage of GPT models as Bi-Encoders remains limited. Rather, there has been interest in using them as generators to produce search training data for encoders \cite{schick2021generating, bonifacio2022inpars}. Previous work has studied the differences in embeddings produced by various language models including BERT and GPT \cite{ethayarajh2019contextual, liu2020survey, cai2020isotropy}. They have found GPT-2 embeddings to underperform on various word embedding benchmarks \cite{ethayarajh2019contextual}. Concurrently to our work, the first trained GPT-based Bi-Encoder, cpt-text, was proposed \citep{neelakantan2022text}. They use a pre-trained decoder and employ two additional training stages, contrastive unsupervised pre-training and supervised fine-tuning. Their models are used for the OpenAI Similarity and Search Embeddings API \cite{openaiembeddings}. Our Bi-Encoders differ from theirs in that we simplify the training process to only fine-tuning, use a novel pooling method and only train bias parameters. cpt-text is most similar to our Bi-Encoders, hence we include their results in our benchmarks.

Cross-Encoders tend to outperform Bi-Encoders \cite{thakur2020augmented}, but are slower as vectors cannot be cached and reused. To balance the trade-offs, multi-stage architectures have been proposed \cite{nogueira2019multi, qu2020rocketqa, khattab2020colbert}. In a two-stage re-ranking setup, the first model processes the entire corpus and the second model is only used on the top $k$ documents returned by the first. In \S\ref{sec:ce}, we use Bi-Encoder (BM25 \cite{robertson2009probabilistic}) + Cross-Encoder re-ranking.

\textbf{Asymmetric Search} means queries and documents are not interchangeable. Finding answers given a question is an asymmetric search problem. Commonly, documents are much longer than queries \cite{thakur2021beir}. We evaluate asymmetric search experiments on BEIR \cite{thakur2021beir}, a recently proposed benchmark consisting of 19 asymmetric search datasets.

\textbf{Symmetric Search} means queries and documents are interchangeable. Finding duplicate questions, where both queries and documents are questions, is a symmetric search problem. We evaluate symmetric search experiments on USEB \cite{wang2021tsdae}, Quora from BEIR \cite{thakur2021beir} and STS-B \cite{cer2017semeval}. In Quora, queries are question titles and documents are question texts. They are often the same with average word lengths of 9.53 and 11.44, respectively \cite{thakur2021beir}. Hence, we consider it more of a symmetric search task. We include Quora in both symmetric and asymmetric experiments.

\section{SGPT Cross-Encoder}\label{sec:ce}

\subsection{Asymmetric Search}\label{sec:ceasym}

\subsubsection{Method}\label{sec:ceasymmeth}

\begin{table*}[t!]
    \small
    \resizebox{\textwidth}{!}{\begin{tabular}{l | c | c c | c c c c | c }
        \toprule
        \multicolumn{1}{l}{\textbf{Training ($\rightarrow$)}} &
        \multicolumn{7}{c|}{Unsupervised} &
        \multicolumn{1}{c}{Unsupervised + Supervised} \\
        \midrule
        \multicolumn{1}{l}{\textbf{Ranking ($\rightarrow$)}} &
        \multicolumn{1}{c|}{Re-rank Top 0} &
        \multicolumn{2}{c|}{Re-rank Top 10} &
        \multicolumn{5}{c}{Re-rank Top 100} \\
        \cmidrule(lr){1-1}
        \cmidrule(lr){2-2}
        \cmidrule(lr){3-4}
        \cmidrule(lr){5-9}
        \multicolumn{1}{l}{\textbf{Model ($\rightarrow$)}}
        & \cite{robertson2009probabilistic}
        & \multicolumn{2}{c|}{SGPT-CE}
        & \multicolumn{2}{c|}{OpenAI Search}
        & \multicolumn{2}{c|}{SGPT-CE}
        & \cite{thakur2021beir} \\
        \multicolumn{1}{l}{\textbf{Dataset ($\downarrow$)}} &
        \multicolumn{1}{c|}{\textbf{BM25}} &
        \multicolumn{1}{c}{\textbf{2.7B}} &
        \multicolumn{1}{c|}{\textbf{6.1B}} &
        \multicolumn{1}{c}{\textbf{Ada}} &
        \multicolumn{1}{c|}{\textbf{Davinci}} &
        \multicolumn{1}{c}{\textbf{2.7B}} &
        \multicolumn{1}{c|}{\textbf{6.1B}} &
        \multicolumn{1}{c}{\textbf{BM25+CE${\clubsuit}$}} \\
        \midrule

   MS MARCO & 0.228
    & 0.249$^\ddagger$ & 0.253$^\ddagger$
    & L & L & 0.278$^\ddagger$ & 0.290$^\ddagger$ & \textbf{0.413}$^\ddagger$ \\  \midrule \midrule
   
    TREC-COVID & 0.688
    & 0.708 & 0.705
    & 0.616 & 0.627 & 0.762 & \textbf{0.791}
    & 0.757 \\
    
    BioASQ & 0.488
    & 0.517 & 0.518
    & L & L & 0.546 & \textbf{0.547} 
    & 0.523 \\
    
    NFCorpus & 0.306
    & 0.319 & 0.323
    & 0.336 & \textbf{0.358} & 0.333 & 0.347
    & 0.350 \\ \midrule
    
    NQ & 0.326
    & 0.358 & 0.366
    & L & L & 0.384 & 0.401
    & \textbf{0.533} \\ 
    
    HotpotQA & 0.602 
    & 0.647 & 0.649
    & L & L & 0.691 & 0.699
    & \textbf{0.707} \\ 
    
    FiQA-2018 & 0.254 
    & 0.305 & 0.313
    & 0.320 & & 0.369 & \textbf{0.401}
    & 0.347 \\ \midrule
   
    Signal-1M (RT) & 0.330
    & \textbf{0.343} & 0.342
    & 0.313 &  & 0.320 & 0.323
    & 0.338 \\ \midrule
    
    TREC-NEWS & 0.405
    & 0.409 & 0.418
    & L & L & 0.434 & \textbf{0.466}
    & 0.431 \\
    
    Robust04 & 0.425
    & 0.438 & 0.444
    & L & L & 0.449 & \textbf{0.480}
    & 0.475 \\ \midrule
    
    ArguAna & \textbf{0.472}
    & 0.379 & 0.376
    & 0.166 & & 0.293 & 0.286
    & 0.311 \\  
    
    Touché-2020 & \textbf{0.347}
    & 0.335 & 0.335
    & 0.332 & & 0.256 & 0.234
    & 0.271 \\ \midrule 
    
    CQADupStack & 0.326 
    & 0.364 & 0.367
    & 0.328 & & 0.405 & \textbf{0.420}
    & 0.370 \\
   
    Quora & 0.808
    & 0.810 & 0.810
    & 0.786 & & 0.792 & 0.794
    & 0.825 \\ \midrule
    
    DBPedia & 0.320
    & 0.341 & 0.340
    & L & L & 0.367 & 0.370
    & \textbf{0.409} \\ \midrule
   
    SCIDOCS & 0.165
    & 0.176 & 0.177
    & 0.161 & & 0.186 & \textbf{0.196}
    & 0.166 \\ \midrule
   
    FEVER & 0.649 
    & 0.706 & 0.723
    & L & L & 0.698 & 0.725
    & \textbf{0.819} \\ 
    
    Climate-FEVER & 0.186
    & 0.179 & 0.189
    & L & L & 0.138 & 0.161
    & \textbf{0.253}\\ 
    
    SciFact & 0.611
    & 0.653 & 0.657
    & \textbf{0.727} & & 0.676 & 0.682
    & 0.688 \\ \midrule
    
    Average & 0.428
    & 0.444 & 0.447
    & & & 0.450 & 0.462
    & \textbf{0.476}\\
    
    Best on & 2
    & 1 & 0
    & 1 & 1 & 0 & \textbf{7}
    & 5\\
   
    \bottomrule
    \end{tabular}}
    \caption{Re-ranking performances on BEIR \cite{thakur2021beir}. OpenAI Search is to be distinguished from the OpenAI Embeddings endpoint. Please refer to Table \ref{tab:beirberesults} in the Bi-Encoder section for a benchmark with the OpenAI Embeddings endpoint. Results on the Search endpoint were produced in October 2021. Scores are \textbf{nDCG@10}. \textit{L}: Dataset is too large for OpenAI's endpoint. $\ddagger$: Used for prompt-tuning (SGPT) or training (BM25+CE). $\clubsuit$: Results from \cite{thakur2021beir}. Other scores are from us. Average scores do not include MS MARCO.}
    \label{tab:beirceresults}
\end{table*}

Given a query $q$, and a document corpus $D$, we are interested in the most likely document $d^*$. Using Bayes' Theorem this can be expressed as:

\begin{equation}
d^* = \arg \max_{d\in D} P(d|q)
= \arg \max_{d\in D} \frac{P(q|d) P(d)}{P(q)}
= \arg \max_{d\in D} P(q|d) P(d)
\end{equation}

Note that $P(q)$ is irrelevant as it is always the same when taking the $\arg \max$ over $D$. Due to variable document lengths and contents it is easier to compare $P(q|d)$ than $P(d|q)$. We hence compute the joint probability of the query tokens $q_{i,..,n}$ given the document tokens embedded in a prompt $P$ as $p(q_i, ..., q_n|p_1,..., p_{i-1})$ ignoring $P(d)$. As long as $P(d)$ does not vary excessively across the corpus $D$, this simplification should produce reasonable scores.

In practice, we use log probabilities \cite{brown2020language, radford2019language}, computed via the log of the softmax of the model output. To have a constant query length $n+1-i$ and avoid abrupt text changes, documents are truncated from the left until the input fits the model's maximum sequence length. We apply these methods to re-rank top $k$ documents returned by BM25 \cite{robertson2009probabilistic}. While re-ranking with BM25 bottlenecks performance, it speeds up experiments. It is not a necessary part of the architecture and therefore not depicted in Figure \ref{fig:sgptgraphic}. 

We experiment with publicly available pre-trained decoder transformers with 125M, 1.3B, 2.7B and 6.1B parameters \cite{gpt-neox, gpt-j}.

\subsubsection{Results}\label{sec:ceasymres}

\begin{table*}[t]
    \centering
    \begin{tabular}{lcccc}
    \toprule
    \textbf{Model Name} & Ada (S) & Babbage (M) & Curie (L) & Davinci (XL)
    \\
    \textbf{Parameters} & 350M (300M) & 1.3B (1.2B) & 6.7B (6B) & 175B (175B)
    \\
    \bottomrule
    \end{tabular}
    \vspace{10pt}
    \caption{OpenAI model parameter estimates. Based on comparing the embedding sizes from the OpenAI docs with the dimensions provided in \cite{brown2020language}. In brackets are numbers for cpt-text models recently provided in \cite{neelakantan2022text}. They differ likely due to removing the language modeling head.}
    \label{tab:oamodels}
\end{table*}

\begin{figure*}[t]
    \centering
    \begin{center}
        \includegraphics[
        width=0.75\textwidth
        ]{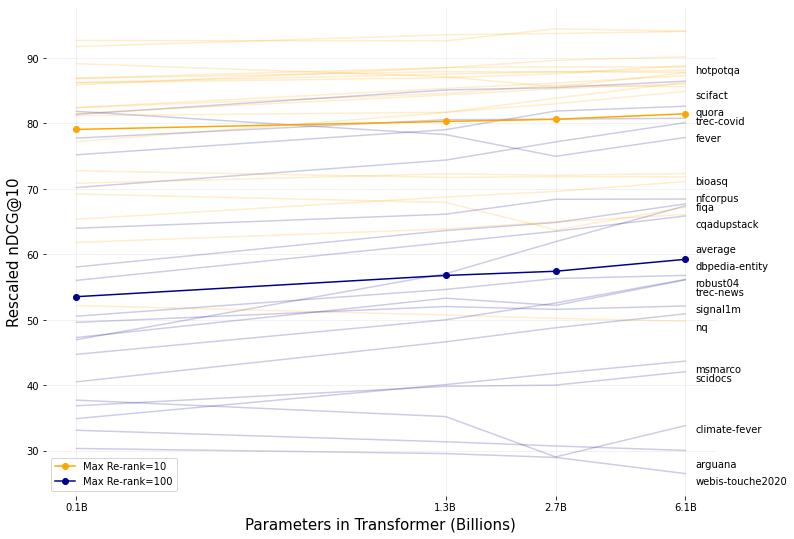}
        \caption{Scaling behavior across parameters and re-ranking for SGPT-CE on BEIR. Scores are rescaled nDCG@10 based on bounds defined in Appendix \S\ref{sec:results}. Dataset labels are ordered by the \textit{Max Re-rank=100} 6.1B performance. The higher on the y-axis, the more bottlenecked is the Cross-Encoder by BM25's performance. Average scores include MS MARCO.}
        \label{fig:sgptscaling}
    \end{center}
\end{figure*}

We perform a search over 12 prompts using the MSMARCO \cite{nguyen2016ms} dataset as provided in BEIR \cite{thakur2021beir}. The prompts and results are in Appendix \S\ref{sec:prompts}. We select the prompt with the best score, $P_G$.

In Table \ref{tab:beirceresults}, we benchmark the resulting SGPT-CE (SGPT-Cross-Encoder). We compare with OpenAI's Search endpoint, which is to be distinguished from their Embeddings endpoint. Please refer to Table \ref{tab:beirberesults} in the Bi-Encoder section for a benchmark with the OpenAI Embeddings endpoint. We provide parameter estimates for the OpenAI model names in Table \ref{tab:oamodels}. We also compare with the current state-of-the-art on BEIR \cite{thakur2021beir}, a BERT-based Cross-Encoder. BM25+CE consists of a pre-trained BERT model that is further fine-tuned on MS-MARCO \cite{nguyen2016ms} in a supervised fashion \cite{thakur2021beir}. SGPT-CE consists solely of the pre-trained GPT model. However, SGPT-CE-6.1B has almost 15x more parameters than BM25+CE significantly increasing latency. In the \textit{Re-rank Top 100} setting, the top 100 documents as returned by BM25 are re-ranked by the respective model. While SGPT-CE-6.1B wins on more datasets than the encoder-based state-of-the-art, its average score is worse. This can be alleviated by not using the same prompt $P_G$ for all datasets. We show in \S\ref{sec:cesym} that SGPT-CE-6.1B can beat BM25+CE on Quora by changing the prompt.

In Figure \ref{fig:sgptscaling}, we investigate how performance scales with model size. As we are in a re-ranking setup, the Cross-Encoder performance is bounded by the documents returned by BM25. We provide the BM25 bounds and additional model results in  Appendix \S\ref{sec:results}. In a \textit{Re-rank Top 10} setting, the model is significantly bottlenecked by BM25. SGPT-CE-6.1B reaches around 80\% of the maximum possible performance. We hence observe high jumps in performance for datasets like HotpotQA \cite{yang2018hotpotqa} or TREC-COVID \cite{voorhees2021trec} as we move to top 100. In fact, the 0.791 nDCG@10 on TREC-COVID in Table \ref{tab:beirceresults} is not possible in a \textit{Re-rank Top 10} setting as the bound is at 0.750. From the results, we infer that performance scales both as we re-rank more documents or increase model size.

\subsection{Symmetric Search}\label{sec:cesym}

We use the same methods outlined in \S\ref{sec:ceasymmeth}, but adapt the prompt for symmetric search. We show this on the example of Quora in Table \ref{tab:quoraablations}. In \S\ref{sec:background}, we have explained why Quora is closer to symmetric search than asymmetric search. We search over several prompts on the smaller 125M parameter model and use the best one on the large model. By doing so, SGPT-CE-6.1B improves by 6\% outperforming all Quora results in Table \ref{tab:beirceresults}. We hypothesize that further customizing the prompt for each dataset could significantly improve performance. However, we highlight that searching prompts for all possible input types may not be feasible in practice and is not considered true few-shot learning \cite{perez2021true}. Hence, the prompt we find for Quora may not generalize well to other symmetric search datasets, a key limitation of this method.

\begin{table*}[t]
    \centering
    \small
    \begin{tabular}{c|p{90mm}|c|c}
    \toprule
    Id & Python & 125M & 6.1B
    \\
    \midrule
    G 
    & Documents are searched to find matches with the same content.\textbackslash nThe document "\{doc\}" is a good search result for "\{query\}
    & 0.764
    & 0.794
    \\
    \midrule
    quoraA 
    & Questions are searched to find matches with the same content.\textbackslash nThe question "\{doc\}" is a good search result for "\{query\} 
    & 0.766
    &
    \\
    \midrule
    quoraB 
    & Below are two similar questions asking the same thing.\textbackslash nThe question "\{doc\}" is similar to "\{query\} 
    & 0.751
    &
    \\
    \midrule
    quoraC
    & These two questions are the same: 1. \{doc\} 2.\{query\} 
    & 0.740
    &
    \\ \midrule
    quoraD
    & Question Body: \{doc\} Question Title:\{query\}
    & 0.782
    & \textbf{0.830}
    \\
    \midrule
    quoraE
    & Question Body: \{shortdoc\} Question Title: \{shortquery\}\textbackslash n Question Body: \{doc\} Question Title: \{query\}
    & 0.773
    &
    \\
    \bottomrule
    \end{tabular}
    \caption{SGPT-CE symmetric search results on Quora. The sum of log probabilities from \{query\} is used as the re-rank score. Overflowing tokens are truncated from the left of \{doc\}. Top 100 documents are re-ranked. Scores are \textbf{nDCG@10}.}
    \label{tab:quoraablations}
\end{table*}

\section{SGPT Bi-Encoder}\label{sec:be}

\subsection{Symmetric Search}\label{sec:bisym}

\subsubsection{Method}\label{sec:bisymmeth}

Like in \S\ref{sec:ceasymmeth}, we first experiment with decoder transformers that have only gone through unsupervised pre-training. In the Bi-Encoder setting, a pooling operation is commonly applied to the model's hidden states to reduce them to a vector whose size is irrespective of sequence length. SBERT \citep{reimers2019sentence} showed that a \textit{MEAN} pooling mechanism outperforms \textit{[CLS]} and \textit{MAX} strategies for a BERT encoder. Due to the causal attention mask in an auto-regressive decoder transformer, tokens do not attend to future tokens like in an encoder transformer. Hence, only the last token has attended to all tokens in a sequence. To account for this information mismatch, we propose to give later tokens a higher weight using a position-weighted mean pooling method:

\begin{equation}
v=\sum_{i=1}^{S}w_ih_i \quad \textrm{where} \quad w_i = \frac{i}{\sum_{i=1}^{S}i}
\end{equation}

where $S$ is the sequence length, $h_i$ the $i$th hidden state and $v$ the query or document embedding. We compare weighted mean pooling with last token pooling, where the hidden state of the final token is the embedding, and regular mean pooling.

We follow recent work \cite{giorgi2020declutr, gao2021simcse,izacard2021towards, neelakantan2022text} and perform supervised contrastive learning with in-batch negatives. Given matching query-doc pairs $\{q^{(i)}, d^{(i)}\}_{i=1}^M$, we optimize the cost function:

\begin{equation}
J_{\mathrm{CL}}(\theta) = \frac{1}{M}\sum_{i=1}^M\log\frac{\exp(\tau \cdot \sigma(f_{\theta}(q^{(i)}), f_{\theta}(d^{(i)})))}{\sum_{j=1}^M\exp(\tau \cdot \sigma(f_{\theta}(q^{(i)}), f_{\theta}(d^{(j)})))}
\end{equation}

where $f_\theta$ is the SGPT model outputting a fixed-size vector, $\sigma$ cosine similarity and $\tau$ a temperature parameter set to $20$ in our experiments. We use GradCache \cite{gao2021scaling} to train with large batch sizes in a limited memory setting. We train on SNLI \cite{bowman2015large} and MNLI \cite{williams2017broad}. We limit the model sequence length to 75 tokens during both training and inference.

We fine-tune only bias parameters and freeze the rest of the model. This has been recently proposed as BitFit \cite{zaken2021bitfit} for BERT encoders. It has been shown to be competitive with full fine-tuning in various scenarios \cite{hu2021lora, wang2021list, logan2021cutting}. Table \ref{tab:bias} shows the number of parameters trained for BitFit models. Due to fewer gradient updates, BitFit significantly reduces GPU memory and time required per step. Further, adding a BitFit checkpoint to an instance with an existing full model will only require storing the different biases. An instance already serving a 22.5GB fp32 GPT-J-6B model requires an additional 22MB of storage to serve an SGPT-5.8B-bitfit model.

\begin{table*}[t]
    \centering\small
    \begin{tabular}{lccccc}
    \toprule
    \textbf{Model Name} & SBERT-Base & SGPT-125M & SGPT-1.3B & SGPT-2.7B & SGPT-5.8B
    \\
    \textbf{Transformer (T.)} & BERT & GPT-Neo & GPT-Neo & GPT-Neo & GPT-J
    \\
    \textbf{Total params} & 109M & 125M & 1.3B & 2.7B & 5.8B
    \\
    \textbf{Bias tensors per T. layer} & 8 & 5 & 5 & 5 & 3
    \\
    \textbf{Bias params} & 103K & 74K & 395K & 658K & 692K
    \\
    \textbf{Bias params \%} & 0.094\% & 0.060\% & 0.030\% & 0.025\% & 0.012\%
    \\
    \bottomrule
    \end{tabular}
    \caption{SGPT parameter overview. Due to the removal of the final language modeling head SGPT-BE-5.8B has 206M parameters less than SGPT-CE-6.1B or GPT-J-6.1B. GPT-Neo models tie the language modeling head weights with the input embeddings, hence there is no parameter difference.}
    \label{tab:bias}
\end{table*}

\subsubsection{Results}\label{sec:bisymres}

\begin{figure*}[t]
    \centering
    \small
    \begin{center}
        \includegraphics[
        width=\textwidth
        ]{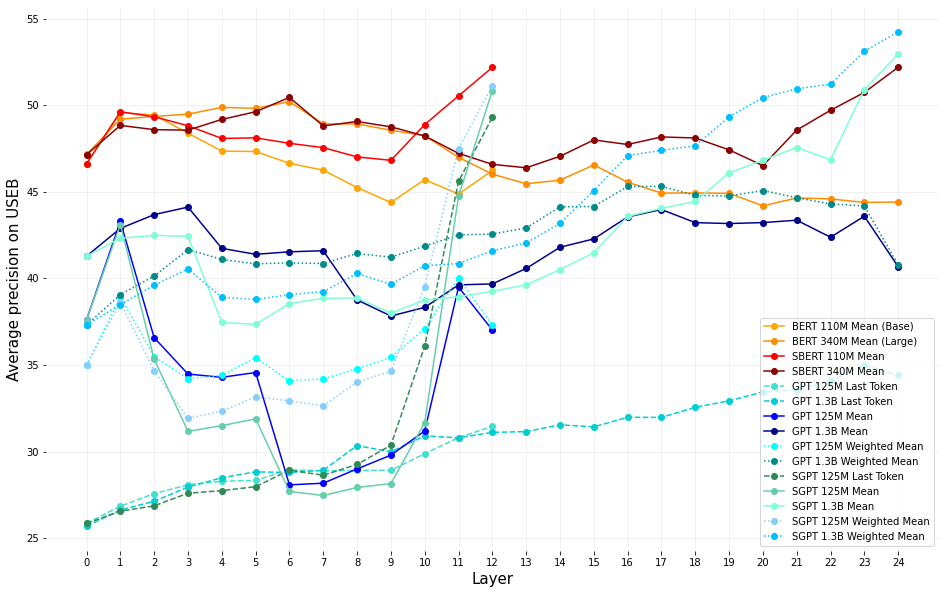}
        \caption{Performance on USEB \cite{wang2021tsdae} by taking the embeddings from certain layers. \textit{S} models are fine-tuned on the same data with the same hyperparameters. Dashed, solid and dotted lines are last token, mean and weighted mean pooling, respectively. Shades of red are transformer encoders, while shades of blue are decoders. The 0th layer is the embeddings prior to the first transformer layer.}
        \label{fig:sgptuseb}
    \end{center}
\end{figure*}

\begin{table*}[t]
    \tiny
    \resizebox{\textwidth}{!}{\begin{tabular}{@{}l|c|c|ccc|ccccc|c|c|c@{}}
    \toprule
    \textbf{Dataset ($\rightarrow$)}
    & \textbf{AskU}
    & \textbf{CQA}
    & \multicolumn{3}{c|}{\textbf{TwitterP}}
    & \multicolumn{5}{c|}{\textbf{SciDocs}}
    & \textbf{Avg}
    & \textbf{Quora}
    & \textbf{STS-B}
    \\
    \textbf{Method ($\downarrow$)}
    &
    &
    & \textbf{TURL}
    & \textbf{PIT}
    & \textbf{Avg}
    & \textbf{Cite}
    & \textbf{CC}
    & \textbf{CR}
    & \textbf{CV}
    & \textbf{Avg}
    &
    & &
    \\
    \hline
    \multicolumn{12}{@{}l@{}}{\emph{OOD Unsupervised}}
    \\
    \hline
    BM25 $\diamondsuit$
    & 53.4
    & 13.3
    & 71.9
    & 70.5
    & 71.2
    & 58.9
    & 61.3
    & 67.3
    & 66.9
    & 63.6
    & 50.4
    & 80.8$^\dagger$
    &
    \\
    \hline
    \multicolumn{12}{@{}l@{}}{\emph{OOD Unsupervised + OOD Supervised (NLI)}}
    \\
    \hline
    SBERT-base-nli-v2-bs64
    & 52.8
    & 11.6
    & 75.5
    & 71.5
    & 73.5
    & 68.0
    & 70.6
    & 71.1
    & 73.5
    & 70.8
    & 52.2
    & 78.9
    & 83.9
    \\
    SBERT-base-nli-v2-bf-bs64
    & 53.8
    & 11.7
    & \textbf{76.6}
    & 72.9
    & 74.8
    & 67.5
    & 70.6
    & 70.8
    & 73.0
    & 70.5
    & 52.7
    & 78.8
    & 81.8
    \\
    SGPT-0.1B-weightedmean-nli-bs64
    & 54.9
    & 11.2
    & 72.7
    & 66.0
    & 69.3
    & 66.2
    & 68.9
    & 68.9
    & 71.7
    & 68.9
    & 51.1
    & 79.5
    & 81.0
    \\
    SGPT-0.1B-weightedmean-nli-bf-bs64
    & 54.9
    & 10.8
    & 72.3
    & 65.3
    & 68.8
    & 64.7
    & 67.4
    & 68.0
    & 70.8
    & 67.8
    & 50.6
    & 77.5
    & 78.6
    \\
    SGPT-0.1B-weightedmean-nli-bf-bs1024
    & 55.7
    & 11.1
    & 72.8
    & 66.5
    & 69.6
    & 65.1
    & 67.8
    & 68.6
    & 70.5
    & 68.0
    & 51.1
    & 79.0
    & 79.5
    \\
    SGPT-1.3B-weightedmean-nli-bf-bs1024
    & 56.0
    & 13.5
    & 75.4
    & 70.7
    & 73.1
    & 70.1
    & 72.9
    & 73.2
    & 75.0
    & 72.8
    & 53.8
    & 82.3
    & 83.9
    \\
    SGPT-2.7B-weightedmean-nli-bf-bs1024
    & 57.5
    & 14.0
    & 75.8
    & 71.0
    & 73.4
    & 72.3
    & 75.4
    & 74.7
    & 76.5
    & 74.7
    & 54.9
    & 82.6
    & 84.7
    \\
    SGPT-5.8B-weightedmean-nli-bf-bs1024
    & 57.1
    & \textbf{16.0}
    & 76.5
    & 76.0
    & 76.3
    & \textbf{75.0}
    & \textbf{78.2}
    & \textbf{77.1}
    & \textbf{78.3}
    & \textbf{77.2}
    & \textbf{56.6}
    & \textbf{84.7}
    & \textbf{85.7}
    \\
    \hline
    \multicolumn{12}{@{}l@{}}{\emph{OOD Unsupervised + OOD Unsupervised + OOD Supervised (NLI, \cite{neelakantan2022text})}}
    \\
    \hline
    Ada Similarity (Mar 2022)
    & 55.9
    & 13.6
    & \textbf{76.6}
    & 76.2
    & 76.4
    & 66.7
    & 71.0
    & 71.3
    & 73.7
    & 70.7
    & 54.1
    & 82.2
    &
    \\
    Curie Similarity (Mar 2022)
    & \textbf{57.3}
    & 15.8
    & 76.3
    & \textbf{78.9}
    & \textbf{77.6}
    & 66.3
    & 71.7
    & 71.3
    & 73.0
    & 70.6
    & 55.3
    & 83.1
    &
    \\
    Davinci Similarity (June 2022)
    & 55.9
    & 
    & 76.3
    & 78.0
    & 77.1
    & 
    &
    &
    &
    &
    &
    &
    &
    \\
    \bottomrule
    \end{tabular}}
    \caption{Results on USEB, Quora and STS-B. Metrics are \textbf{average precision} for USEB, \textbf{nDCG@10} for Quora and \textbf{Spearman} correlation for STS-B. \textbf{bf=BitFit}. \textbf{bs=Batch Size}. \textbf{OOD=Out-of-domain}, to contrast these numbers from in-domain numbers in \cite{wang2021tsdae}. However, fragments may be in-domain due to the large pre-training data of the transformer models. \textit{SGPT-0.1B-weightedmean-nli} performs 2\% worse than \textit{SBERT-base-nli-v2} on USEB, but improves on Quora by 1\%. Note that there is still a size difference of 14\% between the two models. $\diamondsuit$: Results from \cite{wang2021tsdae} except when marked with $\dagger$. CQADupstack and SciDocs differ from the same-name datasets in BEIR.}
    \label{tab:besymres}
\end{table*}

Figure \ref{fig:sgptuseb} shows average precisions on USEB \cite{wang2021tsdae} across different methods and layers. Similar to previous work \cite{ethayarajh2019contextual}, we find that in the unsupervised setting, decoder transformers (GPT) strongly underperform encoders (BERT). However, after fine-tuning on the same dataset with the same hyperparameters, decoders (SGPT) with 125M parameters closely trail the 110M parameter encoder (SBERT) for the 12th layer. Weighted mean pooling outperforms mean and last token pooling for SGPT 125M. When increasing SGPT size ten-fold, the last layer performance (24th layer) increases beyond that of SBERT models. The performance difference of weighted mean pooling compared to mean pooling further widens for SGPT 1.3B.

Table \ref{tab:besymres} provides performance on the individual USEB datasets, Quora and STS-B. STS-B scores should not be the focus of comparison due to the drawbacks highlighted in \cite{wang2021tsdae}. Despite training on less than 0.1\% of parameters BitFit models are within +2 to -2\% of fully fine-tuned ones. BitFit degrades performance more for decoders than encoders. This could be due to the missing bias parameters, see Table \ref{tab:bias}. \cite{zaken2021bitfit} highlights the importance of the query bias vector for BERT, which is not present for SGPT models. \textit{SGPT-5.8B-weightedmean-nli-bitfit} sets an out-of-domain state-of-the-art on USEB, but is outperformed by models trained in-domain in \cite{wang2021tsdae}. We observed performance gains by increasing the training batch size. \textit{SGPT-5.8B-weightedmean-nli-bitfit} is trained with a batch size of 1024. In Appendix \S\ref{sec:results}, we provide results using a lower batch size and additional ablations. Results for Ada and Curie were obtained by querying the OpenAI Similarity Embeddings API in March 2022. They correspond to the cpt-text similarity models from \cite{neelakantan2022text} and we provide their parameters in Table \ref{tab:oamodels}.

\subsection{Asymmetric Search}\label{sec:biasym}

\subsubsection{Method}\label{sec:biasymmeth}

If not otherwise specified, we follow the same setup as in \S\ref{sec:bisymmeth}. For asymmetric search, we train on MS-MARCO \cite{nguyen2016ms}. We limit the model sequence length to 300 tokens during both training and inference. We follow concurrent work \cite{neelakantan2022text} and add enclosing brackets to help the model distinguish between query and document. We embed the tokens of query $q$ in two brackets as $[q_{0-n}]$. For documents, we use curly brackets: $\{d_{0-n}\}$. We add the token ids of the brackets to the already tokenized text to avoid the tokens intermingling. We refer to these special brackets as $specb$.

\subsubsection{Results}\label{sec:biasymres}

\begin{table*}[t!]
    \small
    \resizebox{\textwidth}{!}{\begin{tabular}{l | c c | c | c c c | c c c c }
        \toprule
        \multicolumn{1}{l}{\textbf{Training ($\rightarrow$)}}
        & \multicolumn{2}{c|}{Unsupervised}
        & \multicolumn{1}{c|}{U. + U.}
        & \multicolumn{3}{c|}{Unsupervised + Supervised} 
        & \multicolumn{4}{c}{Unsupervised + Unsupervised + Supervised} \\
        \midrule
        \multicolumn{1}{l}{\textbf{Model ($\rightarrow$)}}
        & \multicolumn{1}{c|}{\cite{robertson2009probabilistic}}
        & \multicolumn{1}{c|}{SGPT-CE}
        & \multicolumn{1}{c|}{\cite{neelakantan2022text}}
        & \multicolumn{1}{c|}{\cite{thakur2021beir}}
        & \multicolumn{1}{c|}{\cite{hofstatter2021efficiently}}
        & \multicolumn{1}{c|}{SGPT-BE} 
        & \multicolumn{1}{c|}{\cite{izacard2021towards}}
        & \multicolumn{1}{c|}{\cite{ni2021large}}
        & \multicolumn{2}{c}{OpenAI Embeddings \cite{neelakantan2022text}} \\
        
        \multicolumn{1}{l}{\textbf{Dataset ($\downarrow$)}} &
        \multicolumn{1}{c|}{\textbf{BM25}} &
        \multicolumn{1}{c|}{\textbf{SGPT-6.1B}} &
        \multicolumn{1}{c|}{\textbf{cpt-text-L${\varheart}$}} &
        \multicolumn{1}{c|}{\textbf{BM25+CE${\clubsuit}$}} &
        \multicolumn{1}{c|}{\textbf{TAS-B${\clubsuit}$}} &
        \multicolumn{1}{c|}{\textbf{SGPT-5.8B}} &
        \multicolumn{1}{c|}{\textbf{Contriever${\spadesuit}$}} &
        \multicolumn{1}{c|}{\textbf{GTR-XXL${\vardiamond}$}} &
        \multicolumn{1}{c}{\textbf{cpt-text-L${\varheart}$}} &
        \multicolumn{1}{c}{\textbf{cpt-text-XL${\varheart}$}} \\
        \midrule

   MS MARCO 
   & 0.228 & 0.290
   & 
   & 0.413$^\ddagger$ & 0.408$^\ddagger$ & 0.399$^\ddagger$
   & 
   & \textbf{0.442}$^\ddagger$
   & & \\  \midrule \midrule
   
    TREC-COVID & 0.688 & 0.791
    & 0.427
    & 0.757 & 0.481 & \textbf{0.873}
    & 0.596 & 0.501 & 0.562 & 0.649 \\
    
    BioASQ & 0.488 & \textbf{0.547}
    &
    & 0.523 & 0.383 & 0.413
    & & 0.324 & & \\
    
    NFCorpus & 0.306 & 0.347
    & 0.369
    & 0.350 & 0.319 & 0.362
    & 0.328 & 0.342 & 0.380 & \textbf{0.407} \\ \midrule
    
    NQ & 0.326 & 0.401
    &
    & 0.533 & 0.463 & 0.524
    & 0.498 & \textbf{0.568} & & \\ 
    
    HotpotQA & 0.602 & 0.699
    & 0.543
    & \textbf{0.707} & 0.584 & 0.593
    & 0.638 & 0.599 & 0.648 & 0.688 \\ 
    
    FiQA-2018 & 0.254 & 0.401
    & 0.397
    & 0.347 & 0.300 & 0.372
    & 0.329 & 0.467 & 0.452 & \textbf{0.512} \\ \midrule
   
    Signal-1M (RT) & 0.330 & 0.323
    & 
    & \textbf{0.338} & 0.289 & 0.267
    & & 0.273 & & \\ \midrule
    
    TREC-NEWS & 0.405 & 0.466
    & 
    & 0.431 & 0.377 & \textbf{0.481}
    & & 0.346 & & \\
    
    Robust04 & 0.425 & 0.480
    & 
    & 0.475 & 0.427 & \textbf{0.514}
    & & 0.506 & & \\ \midrule
    
    ArguAna & 0.472 & 0.286
    & 0.392
    & 0.311 & 0.429 & 0.514
    & 0.446 & \textbf{0.540} & 0.469 & 0.435 \\  
    
    Touché-2020 & \textbf{0.347} & 0.234
    & 0.228
    & 0.271 & 0.162 & 0.254
    & 0.230 & 0.256 & 0.309 & 0.291 \\ \midrule 
    
    CQADupStack & 0.326 & \textbf{0.420}
    & 
    & 0.370 & 0.314 & 0.381
    & 0.345 & 0.399 & & \\
   
    Quora & 0.808 & 0.794
    & 0.687
    & 0.825 & 0.835 & 0.846
    & 0.865 & \textbf{0.892} & 0.677 & 0.638 \\ \midrule
    
    DBPedia & 0.320 & 0.370
    & 0.312
    & 0.409 & 0.384 & 0.399
    & 0.413 & 0.408 & 0.412 & \textbf{0.432} \\ \midrule
   
    SCIDOCS & 0.165 & 0.196
    & 
    & 0.166 & 0.149 & \textbf{0.197}
    & 0.165 & 0.161 & 0.177$^\dagger$ & \\ \midrule
   
    FEVER & 0.649 & 0.725
    & 0.638
    & \textbf{0.819} & 0.700 & 0.783
    & 0.758 & 0.740 & 0.756 & 0.775 \\ 
    
    Climate-FEVER & 0.186 & 0.161
    & 0.161
    & 0.253 & 0.228 & \textbf{0.305}
    & 0.237 & 0.267 & 0.194 & 0.223 \\ 
    
    SciFact & 0.611 & 0.682
    & 0.712
    & 0.688 & 0.643 & 0.747
    & 0.677 & 0.662 & 0.744 & \textbf{0.754} \\ \midrule
   
    Sub-Average & 0.477 & 0.499
    & 0.442
    & 0.520 & 0.460 & \textbf{0.550}
    & 0.502 & 0.516 & 0.509 & 0.528
    \\
    Average & 0.428 & 0.462
    & 
    & 0.476 & 0.395 & \textbf{0.490}
    & & 0.458 & &
    \\
    Best on & 1 & 2
    & 0
    & 3 & 0 & \textbf{5}
    & 0 & 3 & 0 & 4
    \\
    \bottomrule
    \end{tabular}}
    \caption{Comparison of BEIR state-of-the-art models. Keep model size, latency and training time in mind when inspecting this table. Further, this table compares 2 Cross-Encoders and 8 Bi-Encoders, whose respective trade-offs should be considered.  Scores are \textbf{nDCG@10}. $\ddagger$: In-domain performance.  $\clubsuit$: Results from \cite{thakur2021beir}. $\spadesuit$: Results from \cite{izacard2021towards}. $\vardiamond$: Results from \cite{ni2021large}. $\varheart$: Results from \cite{neelakantan2022text} except when marked with $\dagger$. Other scores are from us. Average scores do not include MS MARCO.}
    \label{tab:beirberesults}
\end{table*}

Table \ref{tab:beirberesults} benchmarks \textbf{SGPT-BE-5.8B} (\textit{SGPT-5.8B-weightedmean-msmarco-specb-bitfit}) on BEIR \cite{thakur2021beir} with: (\emph{a}) \textbf{BM25}  \cite{robertson2009probabilistic}, a non-semantic fast baseline (\emph{b}) \textbf{SGPT-CE-6.1B} from \S\ref{sec:ce} (\emph{c}) \textbf{BM25+CE} \cite{thakur2021beir}, the current overall state-of-the-art on BEIR (\emph{d}) \textbf{TAS-B} \cite{hofstatter2021efficiently}, the original Bi-Encoder state-of-the-art on BEIR (\emph{e}) \textbf{Contriever} \cite{izacard2021towards}, a similar training scheme as \cite{neelakantan2022text} but using an encoder transformer
(\emph{f}) \textbf{GTR-XXL} \cite{ni2021large}, the current Bi-Encoder state-of-the-art on BEIR with 4.8 billion parameters using the BERT-like encoder transformer of T5 \cite{raffel2019exploring}
(\emph{g}) \textbf{cpt-text}, a GPT-like decoder transformer architecture concurrently proposed in \cite{neelakantan2022text}. Corresponding parameter estimates are in Table \ref{tab:oamodels}.

SGPT-5.8B achieves the best average nDCG@10 both on the BEIR subset selected in \cite{neelakantan2022text} and on the full BEIR benchmark. It outperforms the roughly same-sized cpt-text-L and the 30x larger cpt-text-XL by 8.1\% and 4.2\%, respectively. Yet, cpt-text models have gone through an additional unsupervised training stage \cite{neelakantan2022text} and are fully trained. SGPT-BE-5.8B fine-tunes just 700K parameters, 0.0004\% of the parameters fine-tuned for cpt-text-XL \cite{neelakantan2022text}. See Table \ref{tab:oamodels} for sizes. We suspect much of the difference to come from the cpt-text model's inferior last token pooling as shown in Figure \ref{fig:sgptuseb}. Further, we suspect that the benefits of the additional unsupervised contrastive pre-training stage diminish when followed by supervised contrastive fine-tuning. SGPT-BE-5.8B improves on the overall state-of-the-art, a Cross-Encoder, by 3\%. It improves on the previously best sentence embeddings (Bi-Encoder) on BEIR, GTR-XXL, by 7\%. However, these improvements come at a significant cost. GTR-XXL has 20\% fewer parameters and its embeddings have 768 dimensions. SGPT-BE-5.8B produces embeddings with 4096 dimensions, hence requiring about 5x more storage. It took the model six days on one Nvidia A100 GPU to encode the entire BioASQ corpus with 15M documents and an average 200 words each \cite{thakur2021beir}. Its comparatively low performance on BioASQ may be improved by increasing the sequence length limit beyond 300, however, requiring additional compute. For SGPT-CE-6.1B, the sequence length limit was 2048 for the combined prompt on all datasets. The high performance on TREC-COVID for SGPT models could be due to the different pre-training datasets. The SGPT pre-training dataset, \textit{The Pile} \cite{gao2020pile}, contains data until mid-2020. This may give the models an information advantage on Covid-19. Lastly, we highlight that on Quora SGPT-BE-5.8B-msmarco is outperformed by SGPT-BE-5.8B-nli from Table \ref{tab:besymres}. Given our classification of Quora as a symmetric search task in \S\ref{sec:background}, this supports our overall distinction between asymmetric and symmetric search. We advise users of our models to classify their tasks as symmetric or asymmetric and use the appropriate model. For non-classifiable embedding tasks, both may work, but we recommend experimenting with embeddings from the symmetric models in \S\ref{sec:bisym} first.

\section{Conclusion and Future Work}\label{sec:conclusion}

This work presented SGPT. Building on SBERT, we proposed modifications to GPT models to use them as Cross- or Bi-Encoders for semantic search.

SGPT-BE uses position-weighted mean pooling and fine-tuning of only bias tensors. At scale, it produces new state-of-the-art sentence embeddings. The model can be used for semantic search or other embedding tasks. We recommend using SGPT-BE-5.8B when compute and storage are of high availability and maximum performance is desired.

SGPT-CE extracts log probabilities of pre-trained GPT models to produce unsupervised state-of-the-art search results. The setup presented can only be used for semantic search. Storage can be limited, but compute should be of high availability for SGPT-CE-6.1B. The prompt and max re-rank parameter can be adjusted depending on performance and latency requirements.

Future research could fine-tune a GPT Cross-Encoder on MSMARCO similar to the BM25+CE model. We suspect that this should outperform the presented non-fine-tuned SGPT-CE model as well as SGPT-BE if enough documents are re-ranked. Further, the combination of SGPT with GPT for generative search results could be interesting. Possibly, SGPT embeddings could be injected into GPT models to generate answers. Lastly, a detailed study of the disadvantages of the missing biases in large GPT models could be helpful to consider their inclusion in the training of future large language models.



\begin{ack}
We thank Constantin Eichenberg and Samuel Weinbach for insightful discussions and valuable feedback throughout the project. We thank Robert Baldock, Marco Bellagente and Koen Oostermeijer for reading drafts of this paper. This work has been supported by OpenAI under the academic access program. 
\end{ack}

\bibliography{custom}
\bibliographystyle{acl_natbib}


\appendix

\clearpage

\section{Additional results}
\label{sec:results}

\begin{table*}[!h]
    \small
    \resizebox{\textwidth}{!}{\begin{tabular}{l | c | c c c | c c c }
        \toprule
        \multicolumn{1}{l}{\textbf{Ranking ($\rightarrow$)}} &
        \multicolumn{1}{c|}{Re-rank Top 0} &
        \multicolumn{3}{c|}{Re-rank Top 10} &
        \multicolumn{3}{c}{Re-rank Top 100} \\
        \cmidrule(lr){1-1}
        \cmidrule(lr){2-2}
        \cmidrule(lr){3-5}
        \cmidrule(lr){6-8}
        \multicolumn{1}{l}{\textbf{Model ($\rightarrow$)}}
        & \cite{robertson2009probabilistic}
        & \multicolumn{2}{c|}{SGPT-CE}
        & \multicolumn{1}{c|}{}
        & \multicolumn{2}{c|}{SGPT-CE}
        & \multicolumn{1}{c}{} \\
        \multicolumn{1}{l}{\textbf{Dataset ($\downarrow$)}} &
        \multicolumn{1}{c|}{\textbf{BM25}} &
        \multicolumn{1}{c}{\textbf{125M}} &
        \multicolumn{1}{c|}{\textbf{1.3B}} &
        \multicolumn{1}{c|}{\textbf{Bound}} &
        \multicolumn{1}{c}{\textbf{125M}} &
        \multicolumn{1}{c|}{\textbf{1.3B}} &
        \multicolumn{1}{c}{\textbf{Bound}} \\
        \midrule

   MS MARCO & 0.228
    & 0.237
    & 0.245
    & 0.383
    & 0.232
    & 0.267
    & 0.664\\  \midrule \midrule
   
    TREC-COVID & 0.688
    & 0.695
    & 0.694
    & 0.750
    & 0.693
    & 0.735
    & 0.988\\
    
    BioASQ & 0.488
    & 0.507
    & 0.514
    & 0.588
    & 0.511
    & 0.528
    & 0.798\\
    
    NFCorpus & 0.306
    & 0.314
    & 0.316
    & 0.364
    & 0.298
    & 0.327
    & 0.513\\ \midrule

    NQ & 0.326
    & 0.336
    & 0.354
    & 0.514
    & 0.319
    & 0.367
    & 0.788\\ 
    
    HotpotQA & 0.602 
    & 0.633
    & 0.645
    & 0.690
    & 0.658
    & 0.688
    & 0.808\\ 
    
    FiQA-2018 & 0.254 
    & 0.281
    & 0.297
    & 0.363
    & 0.280
    & 0.340
    & 0.595\\ \midrule
   
    Signal-1M (RT) & 0.330
    & 0.339
    & 0.343
    & 0.390
    & 0.307
    & 0.322
    & 0.619\\ \midrule
    
    TREC-NEWS & 0.405
    & 0.400
    & 0.402
    & 0.492
    & 0.393
    & 0.443
    & 0.831\\
    
    Robust04 & 0.425
    & 0.419
    & 0.434
    & 0.508
    & 0.382
    & 0.427
    & 0.854\\ \midrule
    
    ArguAna & 0.472
    & 0.394
    & 0.383
    & 0.754
    & 0.315
    & 0.299
    & 0.952\\  
    
    Touché-2020 & 0.347
    & 0.340
    & 0.335
    & 0.467
    & 0.268
    & 0.261
    & 0.881\\ \midrule 
    
    CQADupStack & 0.326 
    & 0.348
    & 0.360
    & 0.426
    & 0.357
    & 0.394
    & 0.637\\
   
    Quora & 0.808
    & 0.794
    & 0.809
    & 0.914
    & 0.764
    & 0.791
    & 0.982\\ \midrule
    
    DBPedia & 0.320
    & 0.328
    & 0.336
    & 0.397
    & 0.329
    & 0.356
    & 0.651\\ \midrule
   
    SCIDOCS & 0.165
    & 0.173
    & 0.177
    & 0.245
    & 0.171
    & 0.185
    & 0.465\\ \midrule
   
    FEVER & 0.649 
    & 0.735
    & 0.718
    & 0.824
    & 0.762
    & 0.729
    & 0.931\\ 
    
    Climate-FEVER & 0.186
    & 0.194
    & 0.191
    & 0.281
    & 0.179
    & 0.167
    & 0.474 \\ 
    
    SciFact & 0.611
    & 0.626
    & 0.645
    & 0.729 
    & 0.621
    & 0.652
    & 0.825\\ \midrule
  
    \textbf{Average} & 0.428
    & 0.436
    & 0.442
    & 0.539
    & 0.423
    & 0.445
    & 0.755\\

    \bottomrule
    \end{tabular}}
    \caption{Additional SGPT Cross-Encoder scores on BEIR. Bounds are the maximum achievable score, given the first-stage BM25 results. We report additional \textit{Max Re-rank=10} scores using OpenAI's search endpoint: TREC-COVID: 0.545 (Ada), 0.539 (Davinci); SciFact: 0.670 (Ada), 0.658 (Davinci). Scores are \textbf{nDCG@10}. Average scores do not include MS MARCO.}
    \label{tab:beirceextra}
\end{table*}

\begin{table*}[!h]
    \small
    \resizebox{\textwidth}{!}{\begin{tabular}{l | c | c c c }
        \toprule
        \multicolumn{1}{l}{\textbf{Model ($\rightarrow$)}}
        & \cite{robertson2009probabilistic}
        & \multicolumn{3}{c}{SGPT-BE} \\
        \multicolumn{1}{l}{\textbf{Dataset ($\downarrow$)}} &
        \multicolumn{1}{c|}{\textbf{BM25}} &
        \multicolumn{1}{c}{\textbf{125M}} &
        \multicolumn{1}{c}{\textbf{1.3B}} &
        \multicolumn{1}{c}{\textbf{2.7B}} \\
        \midrule

   MS MARCO & 0.228
    & 0.279
    & 0.361
    & 0.388 \\  \midrule \midrule
   
    TREC-COVID & 0.688
    & 0.738
    & 0.785
    & 0.807\\
    
    BioASQ & 0.488
    & 0.272
    & 0.347
    & 0.384 \\
    
    NFCorpus & 0.306
    & 0.228
    & 0.321
    & 0.339 \\ \midrule
    
    NQ & 0.326
    & 0.297
    & 0.430
    & 0.467 \\ 
    
    HotpotQA & 0.602 
    & 0.409
    & 0.499
    & 0.528 \\ 
    
    FiQA-2018 & 0.254 
    & 0.211
    & 0.300
    & 0.333 \\ \midrule
   
    Signal-1M (RT) & 0.330
    & 0.236
    & 0.250
    & 0.249 \\ \midrule
    
    TREC-NEWS & 0.405
    & 0.319
    & 0.424
    & 0.438 \\
    
    Robust04 & 0.425
    & 0.313
    & 0.421
    & 0.449 \\ \midrule
    
    ArguAna & 0.472
    & 0.455
    & 0.497 
    & 0.505 \\  
    
    Touché-2020 & 0.347
    & 0.230
    & 0.245
    & 0.235 \\ \midrule 
    
    CQADupStack & 0.326 
    & 0.249
    & 0.320
    & 0.349 \\
   
    Quora & 0.808
    & 0.730
    & 0.853
    & 0.856 \\ \midrule
    
    DBPedia & 0.320
    & 0.227
    & 0.315
    & 0.347 \\ \midrule
   
    SCIDOCS & 0.165
    & 0.121
    & 0.161
    & 0.165 \\ \midrule
   
    FEVER & 0.649 
    & 0.605
    & 0.682
    & 0.728 \\ 
    
    Climate-FEVER & 0.186
    & 0.218
    & 0.266
    & 0.272 \\ 
    
    SciFact & 0.611
    & 0.569
    & 0.683
    & 0.702 \\ \midrule
    
    \textbf{Sub-Average} & 0.477
    & 0.420
    & 0.495
    & 0.514\\
    
    \textbf{Average} & 0.428
    & 0.357
    & 0.433
    & 0.453\\

    \bottomrule
    \end{tabular}}
    \caption{Additional SGPT Bi-Encoder scores on BEIR. Scores are \textbf{nDCG@10}. Average scores do not include MS MARCO.}
    \label{tab:beirceextra}
\end{table*}

\begin{table*}[!h]
    \tiny
    \resizebox{\textwidth}{!}{\begin{tabular}{@{}l|c|c|ccc|ccccc|c|c|c@{}}
    \toprule
    \textbf{Dataset ($\rightarrow$)}
    & \textbf{AskU}
    & \textbf{CQA}
    & \multicolumn{3}{c|}{\textbf{TwitterP}}
    & \multicolumn{5}{c|}{\textbf{SciDocs}}
    & \textbf{Avg}
    & \textbf{Quora}
    & \textbf{STS-B}
    \\
    \textbf{Method ($\downarrow$)}
    &
    &
    & \textbf{TURL}
    & \textbf{PIT}
    & \textbf{Avg}
    & \textbf{Cite}
    & \textbf{CC}
    & \textbf{CR}
    & \textbf{CV}
    & \textbf{Avg}
    &
    & &
    \\
    \hline
    \multicolumn{12}{@{}l@{}}{\emph{OOD Unsupervised}}
    \\
    \hline
    BERT-base-uncased-mean
    & 48.5
    & 6.5
    & 69.1
    & 61.7
    & 65.4
    & 59.4
    & 65.1
    & 65.4
    & 68.6
    & 64.6
    & 46.2
    & 57.3
    &
    \\
    BERT-large-uncased-mean
    & 49.5
    & 6.1
    & 63.2
    & 51.0
    & 57.1
    & 60.7
    & 65.7
    & 65.1
    & 68.6
    & 65.0
    & 44.4
    & &
    \\
    GPT-1.3B-mean
    & 50.9
    & 4.9
    & 56.1
    & 47.5
    & 51.8
    & 50.2
    & 56.8
    & 65.1
    & 59.1
    & 55.1
    & 40.7
    & 45.4
    & 
    \\
    GPT-1.3B-weightedmean
    & 50.2
    & 5.7
    & 50.6
    & 49.1
    & 49.9
    & 53.3
    & 57.0
    & 58.5
    & 61.1
    & 57.5
    & 40.8
    & &
    \\
    GPT-1.3B-lasttoken
    & 45.7
    & 4.5
    & 35.8
    & 34.8
    & 35.3
    & 49.7
    & 52.0
    & 53.2
    & 53.7
    & 52.2
    & 34.4
    & &
    \\
    \hline
    \multicolumn{12}{@{}l@{}}{\emph{OOD Unsupervised + OOD Supervised (NLI)}}
    \\
    \hline
    SBERT-large-nli-v2
    & 53.9
    & 11.8
    & 74.6
    & 70.1
    & 72.3
    & 67.8
    & 70.6
    & 71.8
    & 73.0
    & 70.8
    & 52.2
    & 79.2
    & 84.1
    \\
    SGPT-125M-mean-nli
    & 53.9
    & 10.9
    & 73.7
    & 66.6
    & 70.1
    & 65.9
    & 68.3
    & 68.6
    & 70.9
    & 68.4
    & 50.8
    & 79.1
    & 79.6
    \\
    SGPT-125M-mean-nli-bf
    & 54.9
    & 11.1
    & 72.8
    & 64.3
    & 68.6
    & 63.5
    & 66.6
    & 67.0
    & 69.4
    & 66.6
    & 50.3
    & 77.0
    & 76.9
    \\
    SGPT-125M-mean-nli-linear5
    & 50.1
    & 6.0
    & 67.9
    & 58.4
    & 63.1
    & 53.2
    & 57.1
    & 58.4
    & 61.8
    & 57.6
    & 44.2
    &
    & 62.8
    \\
    SGPT-125M-mean-nli-linearthenpool5
    & 50.1
    & 6.4
    & 69.2
    & 60.3
    & 64.8
    & 55.9
    & 59.5
    & 60.9
    & 64.8
    & 60.3
    & 45.4
    &
    & 68.6
    \\
    SGPT-125M-learntmean-nli
    & 54.1
    & 11.2
    & 73.6
    & 66.8
    & 70.2
    & 65.7
    & 68.4
    & 68.7
    & 71.0
    & 68.4
    & 51.0
    & 79.3
    & 80.1
    \\
    SGPT-1.3B-mean-nli
    & 56.2
    & 12.5
    & 75.5
    & 67.6
    & 71.5
    & 68.6
    & 71.6
    & 72.3
    & 73.7
    & 71.5
    & 53.0
    & 80.5
    & 81.8
    \\
    SGPT-1.3B-weightedmean-nli
    & \textbf{56.8}
    & 13.1
    & \textbf{75.6}
    & 70.1
    & 72.9
    & 71.1
    & 74.6
    & 74.3
    & 76.6
    & 74.2
    & 54.2
    & 81.5
    & 83.1
    \\
    SGPT-1.3B-weightedmean-nli-bf
    & 56.2
    & 12.4
    & 74.5
    & 69.8
    & 72.1
    & 68.3
    & 71.8
    & 72.2
    & 74.0
    & 71.6
    & 53.1
    & 81.9
    & 84.0
    \\
    SGPT-2.7B-weightedmean-nli-bf
    & 56.6
    & 13.2
    & 75.0
    & 72.0
    & 73.5
    & 70.5
    & 73.2
    & 73.1
    & 75.3
    & 73.0
    & 54.1
    & 82.0
    & 85.5
    \\
    SGPT-5.8B-weightedmean-nli-bf-bs48
    & 55.9
    & \textbf{15.0}
    & 74.8
    & \textbf{74.1}
    & \textbf{74.4}
    & \textbf{73.8}
    & \textbf{77.1}
    & \textbf{76.6}
    & \textbf{77.5}
    & \textbf{76.3}
    & \textbf{55.4}
    & \textbf{83.9}
    & \textbf{86.3}
    \\
    \bottomrule
    \end{tabular}}
    \caption{Additional results on USEB, Quora and STS-B. Metrics are \textbf{average precision} for USEB, \textbf{nDCG@10} for Quora and \textbf{Spearman} correlation for STS-B. \textbf{bf=BitFit}. \textbf{bs=batch size}. \textbf{OOD=Out-of-domain}, to contrast these numbers from in-domain numbers in \cite{wang2021tsdae}. However, fragments may be in-domain due to the large pre-training data of the transformer models. CQADupstack and SciDocs differ from the same-name datasets in BEIR.}
    \label{tab:besymresextra}
\end{table*}

\begin{table*}[!h]
    \resizebox{\textwidth}{!}{
    \begin{tabular}{l | c c | c c }
        \toprule
        \multicolumn{1}{l}{\textbf{Model ($\rightarrow$)}}
        & \multicolumn{2}{c|}{OpenAI Embeddings Search}
        & \multicolumn{2}{c}{OpenAI Embeddings Similarity} \\
        
        \multicolumn{1}{l}{\textbf{Dataset ($\downarrow$)}} &
        \multicolumn{1}{c}{\textbf{Ada}} &
        \multicolumn{1}{c}{\textbf{Curie}} &
        \multicolumn{1}{c}{\textbf{Ada}} &
        \multicolumn{1}{c}{\textbf{Curie}} \\
        \midrule

   MS MARCO 
   & 0.37935$^\ddagger$ &
   & & \\  \midrule \midrule
   
    TREC-COVID 
    & 0.68067 & 0.56141 
    & 0.18800 & 0.07612 \\
    
    NFCorpus
    & 0.33170 & 0.38007
    & 0.18166 & 0.19961 \\ \midrule
    
    NQ 
    & 0.42815 &
    & 0.02020 & \\ 
    
    HotpotQA 
    & 0.59393 & 
    & 0.12598 & \\ 
    
    FiQA-2018 
    & 0.38412 & 0.45211 
    & 0.07541 & 0.05138 \\ \midrule
   
    Signal-1M (RT)
    & 0.25388 & 
    & 0.22923 & \\ \midrule
    
    TREC-NEWS 
    & 0.43899 &
    & & \\ \midrule
    
    ArguAna
    & 0.46913 & 0.46978 
    & 0.39647 & \\  
    
    Touché-2020 
    & 0.28679 & 
    & & \\ \midrule 
    
    CQADupStack
    & &
    & 0.10171 & \\
   
    Quora 
    & 0.70572 &
    & 0.82175 & 0.83107 \\ \midrule
    
    DBPedia
    & &
    & 0.03479 & \\ \midrule
   
    SCIDOCS 
    & 0.14827 & 0.17738
    & 0.06275 & \\ \midrule

    SciFact 
    & 0.67255 & 0.74345
    & & 0.46676 \\

    \bottomrule
    \end{tabular}
    }
    \caption{Our results using the OpenAI Embeddings Endpoint in December 2021. They match closely with the results published in \cite{neelakantan2022text}. Scores are \textbf{nDCG@10}. $\ddagger$: In-domain performance.}
    \label{tab:beiroaresults}
\end{table*}

\begin{table*}[!h]
    \resizebox{\textwidth}{!}{\begin{tabular}{@{}l|ccccc|c@{}}
    \toprule
    \textbf{Method}
    & \textbf{NFCorpus}
    & \textbf{FiQA}
    & \textbf{ArguaAna}
    & \textbf{SCIDOCS}
    & \textbf{SciFact}
    & \textbf{Avg}
    \\
    \hline
    \multicolumn{7}{@{}l@{}}{\emph{Unsupervised}}
    \\
    \hline
    BM25
    & 0.30630
    & 0.25407
    & 0.47174
    & 0.16472
    & 0.61100
    & 0.36157
    \\
    \hline
    \multicolumn{7}{@{}l@{}}{\emph{Unsupervised + Supervised (MSMARCO)}}
    \\
    \hline
    SBERT-base-msmarco
    & 0.26316
    & 0.25269
    & 0.43918
    & 0.13302
    & 0.53685
    & 0.32498
    \\
    SBERT-base-msmarco-asym
    & 0.26532
    & 0.22916
    & 0.41694
    & 0.11993
    & 0.49366
    & 0.30500
    \\
    SBERT-base-msmarco-bitfit
    & 0.24934
    & 0.22260
    & 0.46007
    & 0.12843
    & 0.53443
    & 0.31897
    \\
    SGPT-125M-weightedmean-msmarco
    & 0.21628
    & 0.19757
    & 0.41055
    & 0.11720
    & 0.54417
    & 0.29715
    \\
    SGPT-125M-weightedmean-msmarco-asym
    & 0.20919
    & 0.20865
    & 0.35309
    & 0.10618
    & 0.52456
    & 0.28034
    \\
    SGPT-125M-weightedmean-msmarco-bitfit
    & 0.19088
    & 0.17411
    & 0.45265
    & 0.11244
    & 0.53280
    & 0.24381
    \\
    SGPT-125M-weightedmean-msmarco-specb
    & 0.23734
    & 0.22793
    & 0.41137
    & 0.11542
    & 0.58245
    & 0.31490
    \\
    SGPT-125M-weightedmean-msmarco-specb-bitfit
    & 0.22327
    & 0.20911
    & 0.44932
    & 0.11956
    & 0.57703
    & 0.31566
    \\
    SGPT-125M-lasttoken-msmarco-specb
    & 0.14277
    & 0.18626
    & 0.28850
    & 0.11014
    & 0.31202
    & 0.20794
    \\
    SGPT-125M-weightedmean-msmarco-specb-bitfitwte
    & 0.14440
    & 0.09439
    & 0.15428
    & 0.05666
    & 0.32163
    & 0.15427
    \\
    SGPT-1.3B-weightedmean-msmarco-specb-bitfit
    & 0.31748
    & 0.30813
    & 0.47963
    & 0.15827
    & 0.67697
    & 0.38810
    \\
    SGPT-2.7B-weightedmean-msmarco-specb-bitfit
    & 0.32936
    & 0.32469
    & 0.49207
    & 0.16520
    & 0.71228
    & 0.40472
    \\
    SGPT-5.8B-weightedmean-msmarco-specb-bitfit-bs48
    & 0.35330
    & 0.37080
    & 0.49888
    & 0.19389
    & 0.74195
    & 0.43176
    \\
    SGPT-5.8B-weightedmean-msmarco-specb-bitfit-bs256
    & 0.36213
    & 0.37200
    & \textbf{0.51352}
    & \textbf{0.19719}
    & \textbf{0.74693}
    & 0.43835
    \\
    \hline
    \multicolumn{7}{@{}l@{}}{\emph{Unsupervised + Unsupervised + Supervised (MSMARCO)}}
    \\
    \hline
    Ada Search (Dec 2021)
    & 0.33170
    & 0.38412
    & 0.46913
    & 0.14827
    & 0.67255
    & 0.40115
    \\
    Curie Search (Dec 2021)
    & \textbf{0.38007}
    & \textbf{0.45211}
    & 0.46978
    & 0.17738
    & 0.74345
    & \textbf{0.44456}
    \\
    \bottomrule
    \end{tabular}}
    \vspace{10pt}
    \caption{SGPT-BE experiments on a subset of the 5 smallest BEIR datasets by corpus size. The best checkpoint for all models is displayed. \textbf{specb=special brackets}. \textbf{bs=batch size}. \textbf{bitfitwte=BitFit + Word Token Embeddings are trained}. The idea was to help the model learn the special role of the brackets. It did not help. \textbf{asym=Two-tower model} with separate transformers for queries and documents. 
    \textit{SGPT-125M-weightedmean-msmarco-specb} performs 3\% worse than \textit{SBERT-base-msmarco} on average. \textit{SGPT-125M-weightedmean-msmarco-specb-bitfit} performs 1\% worse than \textit{SBERT-base-msmarco-bitfit} on average. Interestingly, Curie beats SGPT-5.8B on this subset, but does not on the bigger subset in Table \ref{tab:beirberesults}. Scores are \textbf{nDCG@10}.}
    \label{tab:beirbesubset}
\end{table*}

\clearpage

\section{Task and Experimental Details}
\label{sec:details}

\subsection{Prompts}
\label{sec:prompts}

\begin{table*}[htbp]
    \scriptsize
    \resizebox{\textwidth}{!}{\begin{tabular}{c|p{55mm}|p{75mm}|c}
    \toprule
    Id & Python & Example & MSMARCO
    \\
    \midrule
    A 
    & \{doc\} \{query\} 
    & If you receive dividends on an investment, those are taxed. Tax on Stocks or ETF's
    & 0.210
    \\
    \midrule
    B 
    & \{doc\}\textbackslash n\{query\} 
    & If you receive dividends on an investment, those are taxed. 
    \newline
    Tax on Stocks or ETF's
    &  0.230
    \\
    \midrule
    C 
    & Document:\textbackslash n\{doc\}\textbackslash n\textbackslash n\newline Query:\textbackslash n\{query\} 
    & 
    Document:\newline
    If you receive dividends on an investment, those are taxed.
    \newline\newline
    Query:\newline
    Tax on Stocks or ETF's
    & 0.264
    \\
    \midrule
    D
    & Body:\textbackslash n\{doc\}\textbackslash n\textbackslash n\newline Title:\textbackslash n\{query\} 
    & 
    Body:\newline
    If you receive dividends on an investment, those are taxed.
    \newline\newline
    Title:\newline
    Tax on Stocks or ETF's
    & 0.242
    \\
    \midrule
    E
    & selected document:\textbackslash n\{doc\}\textbackslash n\textbackslash n\newline relevant query:\textbackslash n\{query\} 
    & selected document:\newline
    If you receive dividends on an investment, those are taxed.
    \newline\newline
    relevant query:\newline
    Tax on Stocks or ETF's 
    & 0.252
    \\
    \midrule
    F
    & The selected text is:\textbackslash n\{doc\}\textbackslash n\textbackslash n\newline The relevant title is:\textbackslash n\{query\}
    & The selected text is:\newline
    If you receive dividends on an investment, those are taxed.
    \newline\newline
    The relevant title is:\newline
    Tax on Stocks or ETF's 
    & 0.246
    \\
    \midrule
    G
    & Documents are searched to find matches with the same content.\textbackslash nThe document "\{doc\}" is a good search result for "\{query\}
    & Documents are searched to find matches with the same content.\newline
    The document "If you receive dividends on an investment, those are taxed." is a good search result for "Tax on Stocks or ETF's
    & \textbf{0.278}
    \\
    \midrule
    H 
    & 
    Documents are searched to find matches with the same content.\textbackslash nDocument: "\{doc\}"\textbackslash n\textbackslash nThe above document is a good match for the query: "\{query\}
    &
    Documents are searched to find matches with the same content.
    \newline
    Document: "If you receive dividends on an investment, those are taxed."
    \newline
    \newline
    The above document is a good match for the query: "Tax on Stocks or ETF's
    &  0.259
    \\
    \midrule
    I 
    & \# Get matching document and query with the same content\textbackslash nget\_document()\textbackslash n\{doc\}\textbackslash nget\_ query\_matching\_document()\textbackslash n\{query\}"
    & \# Get matching document and query with the same content
    \newline
    get\_document()
    \newline
    If you receive dividends on an investment, those are taxed. 
    \newline
    get\_query\_matching\_document()
    \newline
    Tax on Stocks or ETF's
    &  0.253
    \\
    \midrule
    J 
    &
    Documents are searched to find matches with the same content.\textbackslash nThe document "\{shortdoc\}" is a good search result for "\{shortquery\}"\textbackslash nThe document "\{doc\}" is a good search result for "\{query\}
    & 
    Documents are searched to find matches with the same content.
    \newline
    The document "If you receive dividends on an investment, those are taxed." is a good search result for "Tax on Stocks or ETF's"
    \newline
    The document "If you receive dividends on an investment, those are taxed." is a good search result for "Tax on Stocks or ETF's
    & 0.252
    \\
    \midrule
    K 
    & 
    Document:\textbackslash n\{shortdoc\}\textbackslash nQuery:\textbackslash n\{shortquery\}\textbackslash n Document:\textbackslash n\{doc\}\textbackslash nQuery:\textbackslash n\{query\}
    & 
    Document:
    \newline
    If you receive dividends on an investment, those are taxed.
    \newline
    Query:
    \newline
    Tax on Stocks or ETF's
    \newline
    Document:
    \newline
    If you receive dividends on an investment, those are taxed.
    \newline
    Query:
    \newline
    Tax on Stocks or ETF's
    & 0.250
    \\
    \midrule
    L 
    & An intelligent, helpful bot is given. The bot responds "Yes" if the document is a fit to the query and "No" otherwise.\textbackslash n\#\#\#\textbackslash nDocument: \{doc\}\textbackslash nQuery: \{query\}\textbackslash nBot:\{ Yes\}
    & An intelligent, helpful bot is given. The bot responds "Yes" if the document is a fit to the query and "No" otherwise.
    \newline
    \#\#\#
    \newline
    Document: If you receive dividends on an investment, those are taxed. 
    \newline
    Query: Tax on Stocks or ETF's
    \newline
    Bot: Yes
    & 0.112
    \\
    \midrule
    \end{tabular}}
    \caption{Prompts searched over ordered by increasing complexity and their nDCG@10 on MSMARCO using SGPT-CE-2.7B. The example is the shortest query-doc match from FiQA \cite{maia201818}. The sum of log probabilities from \{query\} is used as the re-rank score. Overflowing tokens are truncated from the left of \{doc\}.}
    \label{tab:prompts}
\end{table*}

\subsection{Licenses}
\label{sec:licenses}

Datasets from the BEIR benchmark are licensed under various licenses available in Appendix E of their paper \cite{thakur2021beir}. USEB datasets \cite{wang2021tsdae} are licensed under an Apache 2.0 license.\footnote{https://github.com/UKPLab/useb}. To the best of our knowledge, these datasets do not contain private, personally identifiable information but may contain offensive content. OpenAI models are licensed by OpenAI API to customers via a non-exclusive, non-sublicensable, non-transferable, non-assignable, revocable license.\footnote{https://openai.com/api/policies/terms/}

\subsection{Computational Cost}
\label{sec:compute}

We use the OpenAI API to evaluate Search and Embeddings endpoints. We used tokens equivalent to around 5,000 USD. For SGPT experiments, we use one node with 8x NVIDIA A100 Tensor Core GPU with 40GB memory. For SGPT-CE the evaluation on the entire BEIR suite took around two weeks for the 5.8B model. For SGPT-BE symmetric search training took 21 hours, while asymmetric training took 60 hours for the 5.8B model. Our cluster was provided by Oracle.

\end{document}